# Evaluating Robotic Approach Techniques for the Insertion of a Straight Instrument into a Vitreoretinal Surgery Trocar

Ross Henry[1,2], Martin Huber[1], Anestis Mablekos-Alexiou[1], Carlo Seneci[1], Mohamed Abdelaziz[1,2], Hans Natalius[2], Lyndon da Cruz[2,*], and Christos Bergeles[1,*]

[1]*School of Biomedical Engineering & Imaging Sciences, Kings College London, UK*
[2]*Moorfields Eye Hospital NHS Foundation Trust, London, UK*
*correspondence: ross.henry@kcl.ac.uk, *equal contribution*

## INTRODUCTION

Advances in vitreoretinal surgery have enabled interventions involving precisions previously deemed infeasible [1], with certified systems appearing on the market, e.g. the Preceyes Surgical System offering 20μm accuracy [2]. Despite their benefits, systems add a delay to the interventional workflow, which may be hindering their widespread adoption. A source of delay is the time required to get the system's micro-precise tool into the eye via the Trocar Entry Point (TEP). Automated methods for this task have evaluated a monocular camera feed to facilitate docking [3]. Such approaches may add complexity when seeking regulation. Our paper evaluates the use of a 7-degree-of-freedom (DoF) serial robotic arm to manually position and orient a straight instrument into the TEP. We will compare 3 approaches that use a combination of co-manipulation and teleoperation.

## MATERIALS AND METHODS

The goal is to place a 0.5mm stainless steel rod within a 1 mm custom trocar inserted into the inferior position of a Bioniko Fundus Advanced Eye Phantom [4]. The eye phantom is secured to a table at a height found in theatres.

The rod mimics a variable curvature tool of a prototype Concentric Tube Robot (CTR) when straight building on the work of [5]. It is mounted on a rail within the CTR for the final insertion into the eye and attached to a KUKA LBR Med 7 R800[6]. The LBR is a medically certified 7-DoF serial manipulator, capable of detecting the torques applied at each of its joints, therefore enabling effective human/robot interaction along its body.

A modified surgical microscope foot pedal was used as an input device. The foot pedal featured four buttons, a joystick, and a rocker switch.

The LBR and foot pedal were connected to an Intel-based i7 computer running Ubuntu *22.04* and using ROS 2 Humble for visualisation and data distribution. The LBR was commanded at 100Hz via a ROS 2 integration of the Fast Robot Interface [7]. The LBR joint velocities $dq = KJ^+dx$ were controlled to either facilitate a) a rod tip target velocity dx or b) a simple admittance following $dx = Kf$, with J being the Jacobian, $f = (J^T)^+\tau$ externally applied force-torques, $\tau$ external torques from an inverse dynamics model, and $K$ appropriate gains.

*Trial Methodology*
The trial is comprised 3 tasks: (1) full co-manipulation of the LBR until successful instrument docking, (2) hybrid co-manipulation of the LBR until its end effector is proximal to the TEP, followed by subsequent fine teleoperation of the tool's position and orientation via the foot pedal, and (3) the same hybrid approach with additional assisted guidance using a camera in-built into the system and aligned with the tool, similar to [3]. The task is complete when the participant deems the docking sufficient and extrudes the rod into the phantom via the trocar successfully.

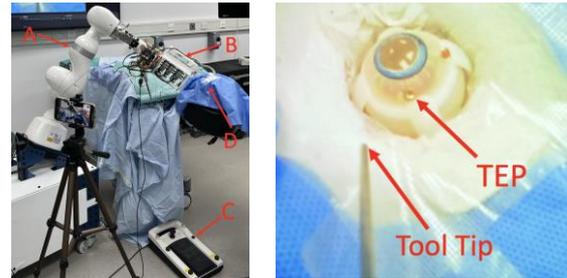

*Figure 1: (Left) Trial setup with components labelled: A) LBR robotic arm, B) CTR, C) modified foot pedal, D) mock head with phantom eye. (Right) View seen from the camera attached to the CTR with the tool tip and TEP highlighted.*

Task 1 required the user to co-manipulate the system to the TEP using the internal torque sensors of the LBR to estimate external forces, $f$, for the admittance controller. The controller is enabled by continually pressing a button on the foot pedal.

For task 2, the system was co-manipulated from the trial start position to approximately 3 cm from the trocar by a member of the engineering team. Afterwards, the participant took over the final approach and teleoperated the end effector to the trocar. The foot pedal provides a 3-DoF velocity control, acting as input for the inverse Jacobian, which can be changed between translational and rotational via the foot pedal. Translational mode moves relative to the tool tip's frame of reference, rotational mode induces a change in orientation while maintaining a centre of rotation about the tip. The participant remained seated throughout the task, mimicking the theatre setup but could look at the phantom head from any angle.

Task 3 followed the same protocol as task 2 but with assistance from a tip-mounted camera. The participant could utilise the camera view via a monitor positioned to the right of them as desired to complete the task.

Each of the tasks was completed a total of 4 times, the first being an introductory run, and the remaining 3 for analysis. A total of 4 participants assisted with the trial, each a vitreoretinal surgeon of varying experience but with no formal training in the use of the proposed system. We recorded each task's time for completion, if the task was successful, the number of collisions of the tool with the eye, and recorded notes on the participants' attempts. An attempt was ultimately deemed unsuccessful if the end effector collided with the cornea or caused excessive deformation to the phantom, as this would translate to possible damage to future patients. A participant could stop and deem their attempt unsuccessful for any reason.

A post-task interview captured the mental demand, physical demand, temporal demand, performance, effort, and frustration via the NASA Task Load Index (NASA TLX). Comments on the control approach were solicited, including a subjective ratio of the time they used the camera view for the final task.

## RESULTS

The average task completion time is summarised in Table 1, the NASA TLX scores are in Figure 2. The co-manipulation approach (Task 1), was consistently faster than the hybrid approaches, with an average completion time of 51.2s compared to 75.6s and 73.8s of Task 2 and 3; a 24.4s and 22.6s improvement respectively.

Task 1, however, had a significantly lower success rate of 42%, with failed attempts caused by LBR joint limit issues, and tool misalignment with the TEP or perceived lag causing excessive deformation of the eye. Task 2's only unsuccessful attempt was due to joint limits during the co-manipulation phase potentially only causing minor surgical delays. Thus, the task had a success rate of 92%.

|        | Average Time (s)   | Success rate (%) |
|--------|--------------------|------------------|
| Task 1 | 51.2 (SD = 23.5)   | 42               |
| Task 2 | 75.6 (SD = 30.7)   | 92               |
| Task 3 | 73.8 (SD = 31.5)   | 100              |

During post-trial interviews, it was noted that there was minimal use of the camera view in Task 3 by the participants with an average camera utilisation of 10%, stating workflow interruptions for low utilisations. Tasks 2 and 3 had lower, thus better scores on the NASA TLX across all dimensions compared to pure co-manipulation (Task 1), as shown in Figure 2. The largest benefits of the hybrid approaches were observed in physical demand and the effort required to complete a task.

Participants showed improvement in task completion time and success rate with each subsequent attempt of Task 2 and 3, indicating a learning curve associated with using the system. It was noted in the post-task interview that the familiarity of the foot pedal assisted with their adoption of the foot pedal control.

## CONCLUSION AND DISCUSSION

This trial successfully evaluated three proposed approaches for the safe and effective placement of a micro-precise tool a vitreoretinal trocar using a medical-grade robotic arm. The pure co-manipulation approach, although the fastest in average completion time, is deemed unfit for patient use due to its high failure rate,

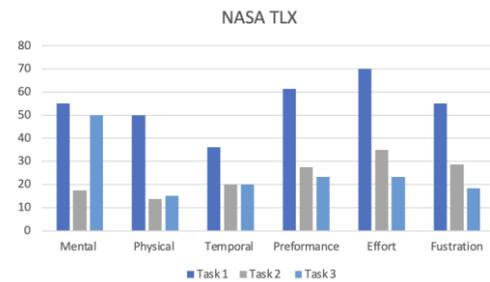

*Figure 2: Averaged NASA TLX results across all participants*

which could potentially cause patient harm, and the increased effort required by the surgeon.

The alternative hybrid methods, combining co-manipulation with teleoperation via a modified foot pedal, completed the tasks in the surgeon's desired 2-minute timeframe while maintaining a near-perfect success rate. Throughout the trial, it was inconclusive which hybrid approach was most beneficial from both the timing and post-task interviews. This may be attributed to the minimal use of the camera view by participants. Therefore, further trials would determine the usefulness of the camera-assisted method, considering the additional complexities it introduces to the system's construction.

## ACKNOWLEDGEMENTS
This work was supported by the National Institute for Health Research (Invention for Innovation, i4i; NIHR202879) and The Sir Joseph Hotung Charitable Settlement. The views expressed are those of the author(s) and not necessarily those of the NHS, the National Institute for Health Research or the Department of Health and Social Care. Acknowledgement to Adriana Namour, Joseph Lovatt-Fraser, and Yawen Xiang for their contribution to creating the camera mount system.